\documentclass[sn-nature,numbers]{sn-jnl}

\usepackage{graphicx}%
\usepackage{multirow}%
\usepackage{amsmath,amssymb,amsfonts}%
\usepackage{amsthm}%
\usepackage{mathrsfs}%
\usepackage[title]{appendix}%
\usepackage{xcolor}%
\usepackage{textcomp}%
\usepackage{manyfoot}%
\usepackage{booktabs}%
\usepackage{algorithm}%
\usepackage{algorithmicx}%
\usepackage{algpseudocode}%
\usepackage{listings}%
\usepackage{lineno}
\usepackage{tcolorbox}
\usepackage{algorithm}
\usepackage{tabularx}
\usepackage{adjustbox}
\usepackage{longtable}
\usepackage{subcaption}
\usepackage{float}

\theoremstyle{thmstyleone}%
\theoremstyle{thmstyletwo}%

\theoremstyle{thmstylethree}%
\def\method{DrugPlayground}

\begin{document}

\title[Article Title]{DrugPlayGround: Benchmarking Large Language Models and Embeddings for Drug Discovery}




\author[1,2,3]{Tianyu Liu}
\equalcont{These authors contributed equally to this work.}
\author[2]{Sihan Jiang}
\equalcont{These authors contributed equally to this work.}
\author[4]{Fan Zhang}
\equalcont{These authors contributed equally to this work.}
\author[5]{Kunyang Sun}

\author[5,6]{Teresa Head-Gordon}
\author*[1,2]{Hongyu Zhao}\email{hongyu.zhao@yale.edu}

\affil[1]{\orgdiv{Interdepartmental Program in Computational Biology \& Bioinformatics}, \orgname{Yale University}, \orgaddress{\country{USA}}}

\affil[2]{\orgdiv{Department of Biostatistics}, \orgname{Yale University}, \orgaddress{\country{USA}}}

\affil[3]{\orgdiv{Broad Institute of MIT and Harvard}, \orgaddress{\country{USA}}}

\affil[4]{\orgdiv{Department of Computer Science and Engineering}, \orgname{The Chinese University of Hong Kong}, \orgaddress{ \city{Hong Kong}}}

\affil[5]{\orgdiv{Pitzer Theory Center and Department of Chemistry}, \orgname{University of California,  Berkeley}, \orgaddress{ Berkeley, CA 94720 \country{USA}}}

\affil[6]{\orgdiv{Department of Bioengineering, Department of Chemical and Biomolecular Engineering}, \orgname{UC Berkeley}, \orgaddress{ Berkeley, CA 94720 \country{USA}}}

\abstract{Large language models (LLMs) are in the ascendancy for research in drug discovery, offering unprecedented opportunities to reshape drug research by accelerating hypothesis generation, optimizing candidate prioritization, and enabling more scalable and cost-effective drug discovery pipelines. However there is currently a lack of objective assessments of LLM performance to ascertain their advantages and limitations over traditional drug discovery platforms. To tackle this emergent problem, we have developed DrugPlayGround, a framework to evaluate and benchmark LLM performance for generating meaningful text-based descriptions of physiochemical drug characteristics, drug synergism, drug-protein interactions, and the physiological response to perturbations introduced by drug molecules. Moreover, DrugPlayGround is designed to work with domain experts to provide detailed explanations for justifying the predictions of LLMs, thereby testing LLMs for chemical and biological reasoning capabilities to push their greater use at the frontier of drug discovery at all of its stages.}


\maketitle

\section{Introduction}
While advances in drug development and clinical application have made substantial contributions to improving human health, these processes remain constrained by scientific, economic, and logistical challenges \cite{dartois2022anti, mak2024artificial, miao2024ai}. In particular, the escalating cost  associated with \textit{de novo} drug design or drug repurposing and the prolonged timelines for experimental validation underscore the need for more efficient paradigms \cite{niazi2025artificial}. 

Recently, Large Language Models (LLMs) have emerged as transformative tools in biology and medicine \cite{chakraborty2023artificial, pal2023chatgpt, tian2024opportunities}, enabling a wide range of applications in drug discovery\cite{Cavanagh2026}, drug development and optimization\cite{Lu2025}, and therapeutic repurposing of existing drugs\cite{Yan2024,More2026}. LLMs are known to be a special class of foundation models (FMs) \cite{bommasani2021opportunities, naveed2023comprehensive,Yuan2026}, trained with a large-scale natural language corpus and capable of performing various text-related downstream applications. LLMs have been introduced into biomedical research to handle patient data such as clinical reports or to serve as lab assistants for summarizing research manuscripts or providing data analysis. Researchers have also adapted or developed LLMs and molecular foundation models (MFMs)\cite{cui2025towards} for generating novel drug molecules\cite{Cavanagh2026} or fine-tuning them as protein language models (PLMs) \cite{ofer2021language} for generating protein sequences with new function. For example, Chemcrow\cite{m2024augmenting} leverages LLMs and several chemistry-related tools to augment the capacities of LLMs for solving molecular-related tasks in an automated pipeline. SmileyLlama explores the chemical space of drug-like molecules through optimized engineered prompts of SMILES strings, and in turn can be further optimized to favor ligand poses with good binding affinity when put into a reinforcement learning context.\cite{Li2024,Cavanagh2026}. SynLlama finetunes an LLM to generate full synthetic pathways of drug molecules made of commonly accessible organic building blocks and robust reaction templates as the underlying learned syntax.\cite{Sun2025} DrugAgent attempts to discover new drugs based on a multi-agent collaboration\cite{liu2024drugagent} while DTI-LM \cite{ahmed2024dti} designs a language model-powered framework to predict drug target interaction. Several tools, including BAITSAO \cite{liu2025building}, CancerGPT \cite{li2024cancergpt}, and SynerGPT \cite{edwards2023synergpt} utilize LLMs and/or their generated embeddings to predict when drugs show a synergistic effect. 

These examples illustrate the power of LLM-based tools to facilitate drug discovery, however, there are also concerns about the safe application of LLMs in the pharmaceutical field.\cite{murakumo2023llm} Murakumo and co-workers have pointed out that LLM-assisted drug design can only handle molecules with a relatively simple chemical structure, or hallucinate unrealistic chemistry, and thus cannot be generalized to meaningfully broaden the chemical space exploration for novel and viable drug candidates. LLMs are also vulnerable to various types of medical misinformation that arise through indiscriminate digestion of natural language data that is "poisonous" in suggesting medications that threaten patient safety\cite{alber2025medical}. Moreover, early-stage exploration shows that LLMs do not always outperform trained from scratch deep learning models in predicting drug properties or binding sites. These issues create uncertainty and dim the possibly bright future prospects for powerful utilization of LLMs for drug discovery pipelines. Therefore there is an urgent need to develop a benchmark platform to evaluate LLMs and gain new insights to better assess the future of LLMs for drug-associated applications.

Here, we introduce \method{}, a unified benchmarking platform designed to systematically evaluate the utility of LLMs for drug discovery and related applications across four representative tasks: drug function analysis to ascertain the pharmacological character that explains drug potency, drug-target interaction prediction such as binding affinities, prediction of synergistic drug combinations with greater therapeutic efficacy, and drug perturbation prediction of cellular responses to drug molecules. Collectively, these tasks span key stages of the research and development (R\&D) drug discovery pipeline and play critical roles in determining the effectiveness and efficiency of therapeutic interventions. Leveraging these benchmarks, we comprehensively assess the capacity of LLMs to learn meaningful drug representations and to generalize knowledge across diverse compounds and biological contexts. Beyond its comprehensive coverage, \method{} is inherently flexible and extensible, enabling the evaluation and development of emerging LLM-based methodologies for drug analysis. Finally, through detailed error analyses and case studies, we elucidate the strengths and limitations of current LLM approaches, providing actionable insights to guide future model and benchmark design that balances accuracy, cost, and runtime.  

\section{Results}
\subsection{Overview of \method{}} 
\begin{figure}[H]
    \centering
    \includegraphics[width=0.9\linewidth]{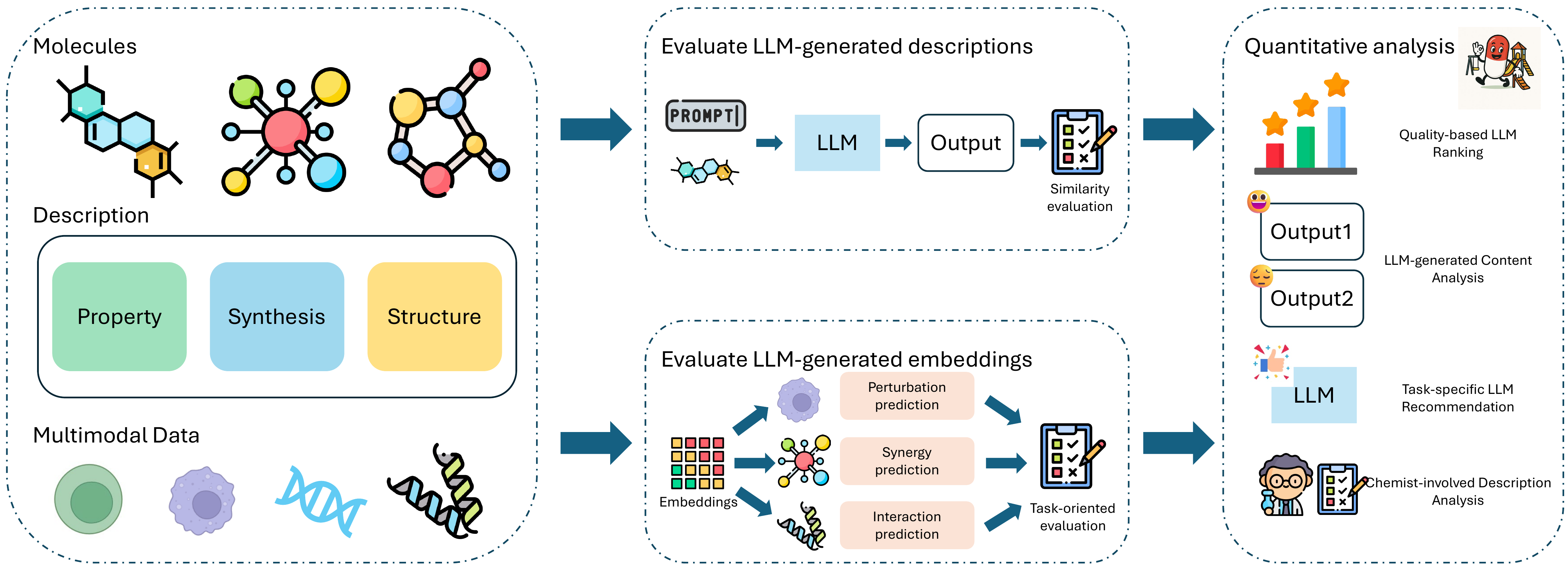}
    
    \caption{\textit{Overview of \method{}}. We prepare datasets with molecule-text-paired information as well as other multimodal sources. Our benchmark contains two different branches, one is for evaluating LLM-generated textual content, the other is for evaluating LLM-generated embeddings. Our evaluation and quantitative analysis not only includes numerical metrics, but also feedback from chemists. We also provide recommendations to choose the most suitable setting for different tasks.}
    \label{fig:overall}
\end{figure}

\method{} is a versatile benchmarking platform built on paired datasets to systematically evaluate the strengths and weaknesses of LLMs for drug-centered analyses. Figure \ref{fig:overall} displays an overview of the designed datasets and benchmarking framework.   
\noindent
We assess LLM capabilities across two complementary tasks: (i) description-based evaluations of drug function analysis, and (ii) drug embedding–based evaluations, including similarity analysis, drug synergy prediction, drug–target prediction, and perturbation outcome prediction. Drug descriptions are generated by advanced LLMs under diverse prompt formulations, and the corresponding drug embeddings are derived using the embedding models of different LLMs applied to these descriptions. For each task, we carefully curate datasets to ensure fair comparisons and to rigorously prevent data leakage.   

\subsection{Evaluation of LLM-generated text for drug properties} 
We first examine whether LLM-generated content is sufficiently precise to accurately summarize established properties of selected drugs. To this end, we collect drug names and their corresponding textual descriptions from the large-scale molecular–text database MolTextNet \cite{zhu2025moltextnet}, which serves as the ground-truth reference. We then compute both schema-driven and structure-driven metrics to quantitatively compare LLM-generated descriptions, produced under different prompt settings, against the ground-truth descriptions. The five LLMs tested are Claude-sonnet-20250514 (Claude-sonnet4) \cite{anthropic_claude4}, DeepSeek-v3 \cite{ds_v31}, GPT-4o \cite{oai_gpt4o}, Gemini-1.5-pro (Gemini-1.5P) \cite{google_gemini}, and Mistral-large-2411 (Mistral-large) \cite{mistral_large}. We examine each LLM evaluated at six temperature settings under different prompting conditions, and compute the average BERT, ROUGE-1, ROUGE-2, ROUGE-L, and BLEU scores, and summarized in Supplementary Figure S1. In addition, we quantify overall performance using a Normalized Total score, defined as the mean of the five individually normalized metrics.

\begin{figure}[h]
    \centering
    \adjustbox{center}{
        \includegraphics[width=1\linewidth]{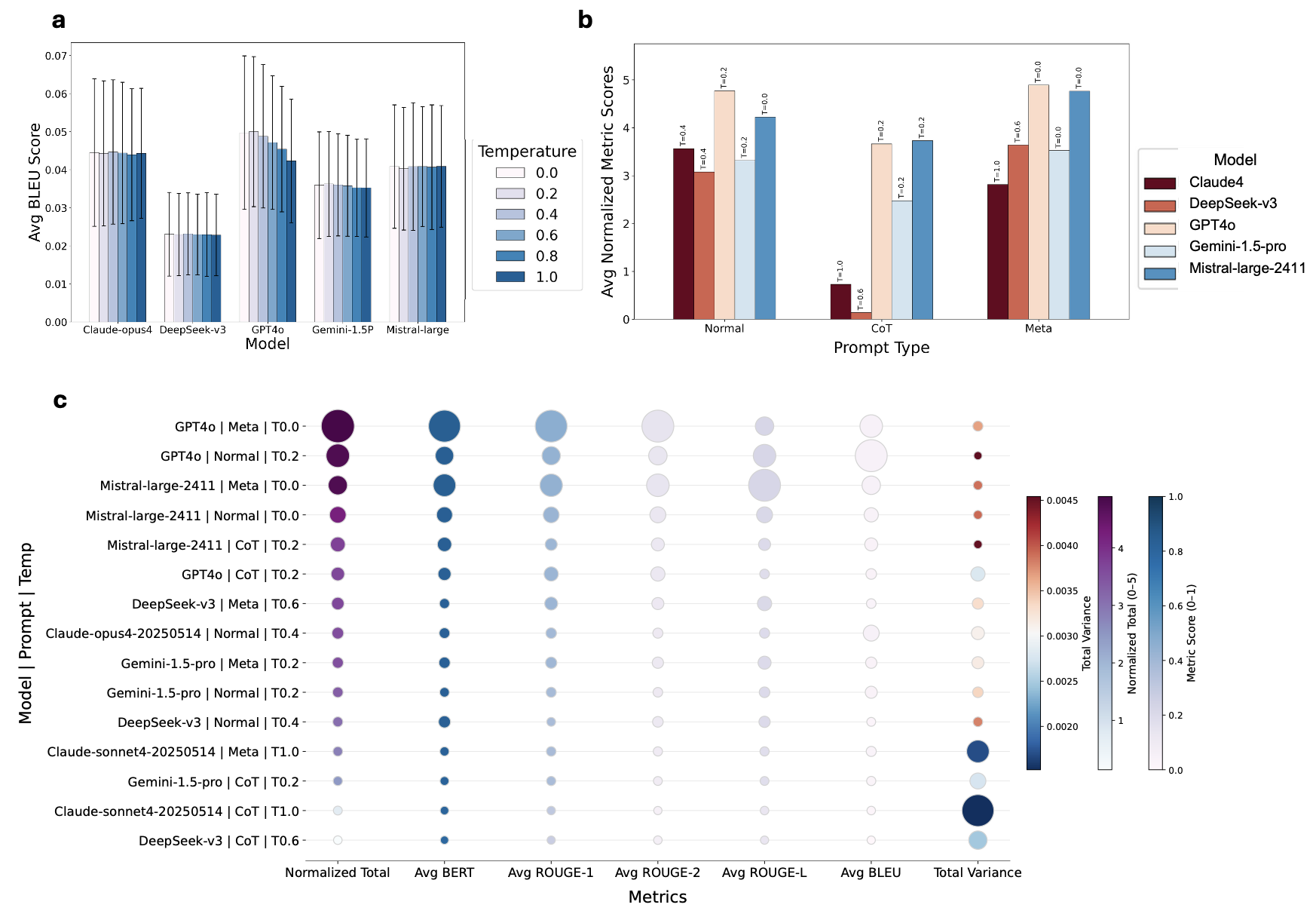}}
    \caption{\textit{Model Performance in Terms of Text Generation} (a) Five LLMs' BLEU scores under standard prompts across temperature. (b) The best performance achieved by each LLM at the optimized temperature across different prompts: standard, chain of thought (CoT), and meta-cognition (Meta) prompts. (c) The five evaluation metrics on their original scales of 0-1, with combinations ordered along the y-axis in descending order of their average normalized metric scores.  The Normalized Total has a 0-5 scale. The total variance is the sum of across-drug variances, which were computed by the variance between the mean metric scores of different drugs' descriptions. We consider three different prompting approaches, including normal, domain-specific meta prompting (Meta), and Chain-of-Thought (CoT). }
    \label{fig: text}
\end{figure}

Figure \ref{fig: text} establishes that temperature tuning exerts a systematic yet model-dependent influence on performance. For most models, lower temperature settings consistently improve most of the metrics, whereas higher temperatures introduce increased variability without yielding consistent gains in generation quality. While semantic similarity measured by BERT remains relatively stable across temperature settings, with variations typically below 0.15\%, Figure \ref{fig: text}(a) demonstrates that each LLM exhibits a distinct optimal temperature regime: GPT-4o and Gemini-1.5-pro achieve their best performance at lower temperatures, Mistral-large-2411 favors the lowest temperature setting, while Claude-sonnet4 and DeepSeek-v3 peak at moderately higher temperatures. Across all evaluated LLMs and temperature settings, GPT-4o consistently demonstrates the strongest overall performance, achieving the highest aggregated scores across all five evaluation metrics with a standardized prompt as seen in Figure \ref{fig: text}(b). Its performance gains are substantial rather than marginal, indicating a clear and consistent advantage in generation quality over the other models. Mistral-large-2411 emerges as the closest competitor, showing particularly strong performance on ROUGE-based metrics as seen in Figure \ref{fig: text}(c), although its overall scores remain lower than those of GPT-4o. Claude-opus4-20250514 and Gemini-1.5-pro form a middle-performance tier, whereas DeepSeek-v3 consistently ranks lowest across all evaluated metrics.

Beyond standard prompting, we further evaluate model performance under chain of thought (CoT) and meta-cognition (Meta) prompts to assess the impact of prompt design on generation quality.  Table \ref{tab: prompts} in the Methods section illustrates the nature of the three types of prompts  that we assess in Figure \ref{fig: text}(b). In this case we evaluate each of the LLMs at their corresponding optimal temperature derived from standard prompting, and consider the  Normalized Total score as our performance metric. Overall, prompt selection introduces clear and systematic performance differences across models (also see Supplementary Figure S1). We find that Meta prompting generally yields the highest average normalized scores, consistently improving upon performance achieved with standard prompts, whereas all five LLMs attain much lower scores when using CoT prompts to generate drug descriptions. Hence Meta prompts more effectively guide LLMs toward high-quality, reference-aligned drug descriptions, wheras CoT prompts tend to introduce redundant phrasing or explicit reasoning artifacts, which reduces lexical and structural alignment with the reference texts. Figure \ref{fig: text}(c) also provides a detailed ranking of the top 15 model–prompt–temperature combinations across all five metrics, further reinforcing the conclusions shown in Figure \ref{fig: text}(b).

Beyond average trends, the contrast between the top- and bottom-ranked model–prompt–temperature configurations highlights the magnitude of performance variability induced by configuration choices. Under the standard prompting, GPT-4o already produces substantially higher-quality descriptions than DeepSeek-v3 across all evaluation metrics, as shown in Supplementary Table S\ref{tab: worst-vs-best}. With alternative prompting strategies, this gap further widens: GPT-4o paired with the Meta prompt achieves the highest overall performance, whereas DeepSeek-v3 with the CoT prompt ranks lowest. The top tier is dominated by GPT-4o and Mistral-large-2411, which together account for the majority of high-ranking configurations across prompt types. This concentration indicates that the benefits of prompt engineering are amplified when applied to strong base models, rather than serving as a mechanism to compensate for weaker ones. 

Figure \ref{fig: text}(c) also provides specific analyses of LLM output variance, in which the prompting strategy has a clear and systematic impact on robustness, with CoT prompting consistently yielding  the lowest across-drug variance, indicating more uniform behavior across different drugs, whereas standard prompting exhibits the highest variance and is therefore the least stable. Meta prompting falls between these two extremes, providing a trade-off between improved average performance and robustness. These patterns suggest that explicitly structured reasoning can act as a regularizer on model outputs, even when it does not maximize absolute performance scores. Temperature effects on stability are also prompt-dependent rather than monotonic. Low temperature alone does not guarantee robustness, as unstable behavior can still emerge under weak prompt designs. In contrast, moderate-to-high temperatures can remain stable when paired with, for example, CoT prompts. Within individual models, switching from suboptimal to optimal prompting strategies leads to large reductions in variance for most models, reinforcing the conclusion that prompt design outweighs temperature choice in determining stability. Mistral-large-2411 is an exception, as CoT does not improve its stability. 

To further characterize variability, we also define within variance as the mean of the variance of the three samples for each of the 862 drugs, and the total variance consists of the sum of across-drug and within-variance shown in Supplementary Figure \ref{fig: variances}. Across all configurations, total variance remains relatively small, indicating that the selected ``best" configurations are generally stable. Overall, the Claude-family models exhibit the strongest across-drug stability, while GPT4o achieves the highest performance at the cost of increased variability. This highlights a trade-off between peak performance and robustness across heterogeneous drug descriptions.

We also find that some LLMs' outputs suffer from  inconsistency, incompleteness, truncation, and hallucinations. The reference descriptions in the sample dataset extracted from the MolTextNet dataset \cite{zhu2025moltextnet} usually starts with a drug name, for example, ``1-phenyl-3-(3,4,5-trimethoxyphenyl)urea", with the canonical SMILES string ``...". Or it starts with ``the compound named \textit{US9096596}, 36 has a canonical SMILES representation of ...". Supplementary Table S\ref{tab: ITH} provides representative examples of problematic responses generated under different model–prompt–temperature combinations. With respect to inconsistency, nearly all three sampled descriptions for each drug exhibit minor variations, regardless of model, prompt type, or temperature. Even at zero temperature, the overall meaning and most of the content remain consistent, with only slight lexical differences observed. Regarding truncation, this issue appears across all three prompting strategies but occurs most frequently in descriptions generated using CoT prompts.

Hallucinations are observed predominantly in responses generated using CoT prompts, often manifesting as unnecessary sentences that explicitly reflect the model’s intermediate reasoning process. Another common form of hallucination involves discrepancies between specific numerical or factual attributes in the generated descriptions and those in the ground truth. For instance, the molecular weight of ``Cbz-Arg-Leu-Sta-NHcHex" is 659.87 in the ground truth. However, in one of the three sample descriptions generated from GPT4o using the Meta prompt at temperature 0.0 tells that its molecular weight is 701.95. Beyond numerical discrepancies, such errors also manifest as systematic biases in chemical and pharmacological properties. For instance, generated descriptions occasionally report incorrect molecular formulas, functional groups, or stereochemical characteristics, even when compound names are correctly identified. These errors suggest that the LLM may rely on approximate pattern recognition rather than precise chemical foundations, particularly for less common or structurally complex drugs. Additional hallucination events involve  composite identity drift, rule-based reasoning errors, quantitative omissions, and overgeneralized pharmacological assertions. 

One thing that is noteworthy to mention here is that descriptions generated by the Mistral-large-2411 model with the normal prompt template frequently include structured chemical formulas or reaction-style representations, which differ from the paragraph-based or bullet-point summaries produced by other LLM configurations. While this structured format may create an impression of technical rigor, it does not necessarily correspond to higher factual accuracy and may still contain incorrect or incomplete chemical information, as shown in Supplementary Figure \ref{fig:failure_examples}.

\subsection{Evaluation of LLM-generated embeddings for diverse drug applications}
Given the central role of drug embeddings in scientific research \cite{karim2019drug, mohamed2020discovering}, particularly in AI-driven drug discovery, we next turn to a systematic evaluation of drug embeddings derived from different LLMs across multiple downstream tasks, including generating meaningful embeddings for drug representation, and predicting drug synergism, drug-target interaction, and the physiological response to perturbations introduced by drug molecules.

\begin{figure}[h]
    \centering
    \adjustbox{center}{\includegraphics[width=1\linewidth]{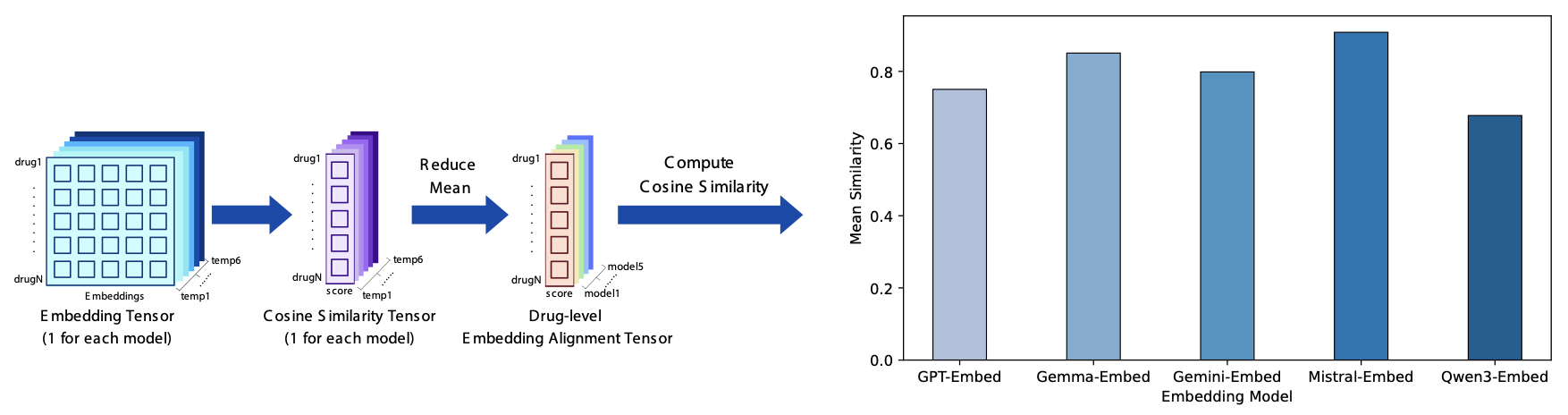}}
    \caption{\textit{Model Performance in Terms of Embedding for Drug Representation} (a) The workflow of evaluating the embeddings generated from the best descriptions evaluated using GPT4o with a Meta prompt at temperature 0.0. Average cosine similarity of generated embeddings from five LLMs across temperatures. Each bar shows the average cosine similarity between the embeddings from generated content and embeddings from the ground truth based on a certain model at the corresponding temperature.}
    \label{fig: embed}
\end{figure}

\textbf{Evaluations of LLM-generated embeddings for drug representations.} Previous research has shown that LLMs can learn representations of languages by compressing the information from text space to generate an embedding space. Such embeddings can help with several downstream applications, thus we also evaluate the quality of embeddings from various LLMs in representing a drug as illustrated in Figure \ref{fig: embed}(a). To perform this evaluation, we select the best LLM model combination from Figure \ref{fig: text}(b), which is GPT4o using Meta prompt at temperature 0.0, to generate text descriptions using text-embedding-3-large (GPT-Emb) \cite{oai_gpt4o}, Gemma-300m (Gemma-Emb) \cite{google_gemma}, gemini-embedding-exp-03-07 (Gemini-Emb) \cite{google_gemini}, mistral-embed (Mistral-Emb) \cite{mistral_large}, Qwen3-Embedding-8B (Qwen3-Emb) \cite{ali_qwen3} . We evaluate the cosine similarity between embeddings from LLM-generated text and ground truth text from MolTextNet \cite{zhu2025moltextnet}, in which a higher cosine similarity represents a better embedding model.

Moreover, to assess the consistency of different language models in capturing drug semantics, Figure \ref{fig: embed}(b) assesses drug-level cosine similarity scores across models. For each model, we first calculated the cosine similarity between the embedding of a drug description generated at each temperature and the corresponding ground-truth embedding of the same drug. These similarities were then averaged across temperatures, yielding a single similarity score per drug for each model. According to Figure \ref{fig: embed}(b), all embedding models can generate high-quality representations reflected by high cosine similarity (larger than 0.7), except Qwen3-Emb. The highest performer is Mistral-Emb, followed by Gemma-Emb and Gemini-Emb, indicating that the power of embedding models in expressing the textual similarity is not closely relevant to parameter scale. 


\textbf{Evaluations of LLM embeddings for drug synergy prediction.} To provide a more comprehensive approach for evaluating the quality of LLM embedding layers, we conducted analysis in conjunction with practical application scenarios. We next evaluate the use of drug embeddings for predicting drug synergy. Drug synergy is defined as a therapeutic effect in which the combined use of two drugs for a given disease exceeds the effect of either drug alone. Prior work, such as BAITSAO \cite{liu2025building}, has demonstrated that LLM-derived drug embeddings can outperform traditional approaches in drug synergy prediction; however, the relative merits of embeddings from different LLMs remain underexplored. To address this gap, we select established datasets containing drug information, synergy annotations, and cell line contexts \cite{preuer2018deepsynergy,el2023marsy}, and conduct a comprehensive benchmarking analysis of drug synergy prediction based on LLM embeddings within the BAITSAO framework. Using this setup, we train both classification and regression models to predict and evaluate synergy effects, and further collaborate with chemists to interpret the biological and chemical insights derived from these experiments.

\begin{figure}[ht]
    \centering
    \includegraphics[width=1\linewidth]{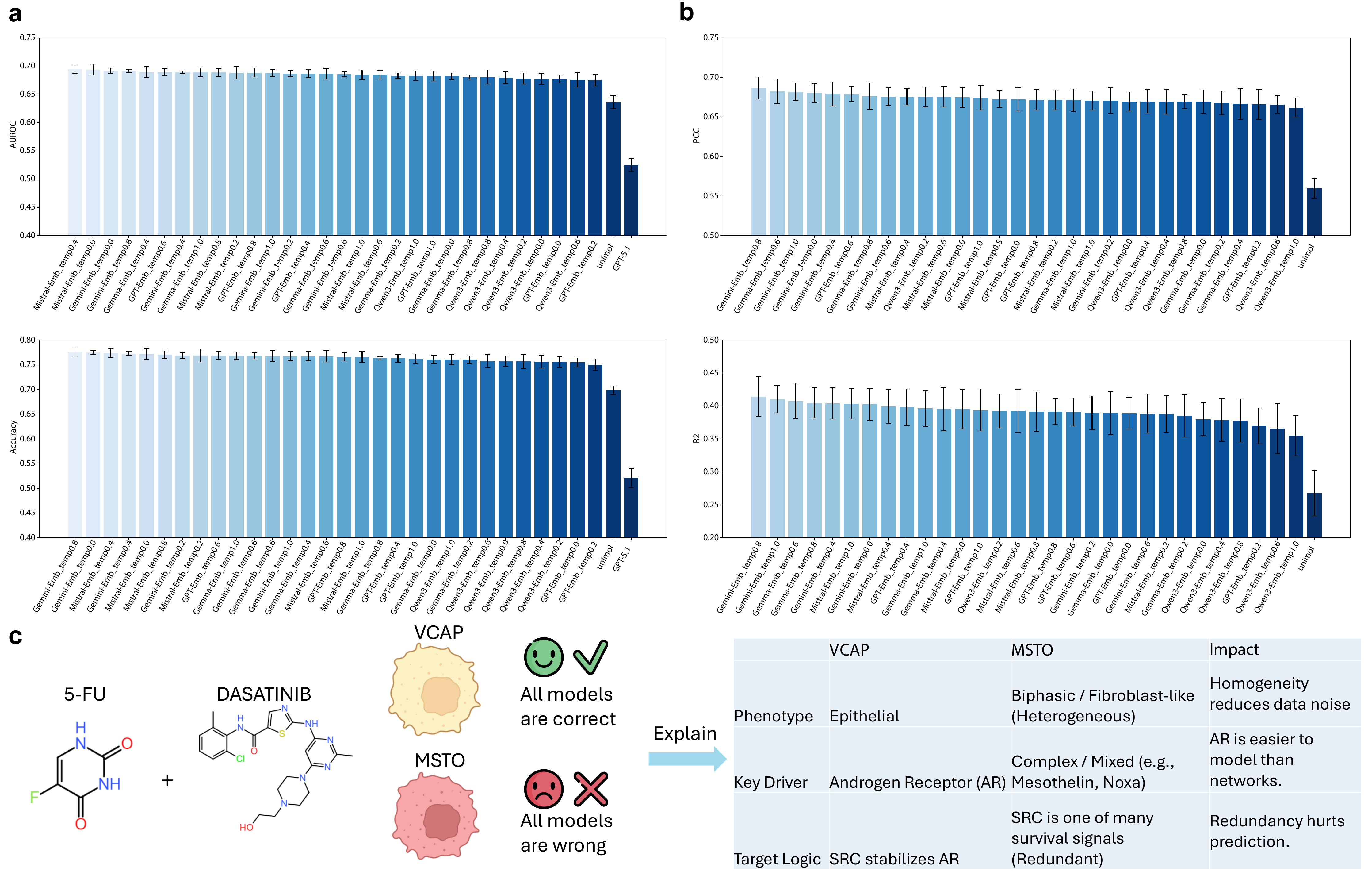}
    \caption{Benchmarking results of drug synergy prediction. (a) AUROC and accuracy across all benchmark methods for the classification task. (a) PCC and R2 across all benchmark methods for the classification task. (c) Exploration of shared capacities across different LLM-generated embeddings from the perspective of biology and chemistry.}
    \label{fig:syner pred}
\end{figure}

Figures \ref{fig:syner pred}(a) and (b) demonstrate the overall superior performance of using LLM-derived embeddings to represent drugs and cell lines for drug synergy prediction, evaluated under both classification (a) and regression (b) settings. We compare this approach against molecular foundation model (MFM) embeddings derived from Uni-Mol \cite{zhou2023unimol}, as well as direct LLM inference–based prediction using an advanced LLM, GPT-5.1 \cite{OpenAI_GPT5_SystemCard}. In the latter case, the task is formulated as a question–answering problem that outputs a probability of synergy and is therefore applicable only to the classification setting. Neither MFM embeddings nor inference-based prediction outperforms the LLM-embedding-based approach. Among the top-performing models, Gemini-Emb and Mistral-Emb achieve the highest average performance across tasks, highlighting their strong capability to capture drug interaction–relevant representations, and constitute promising starting points for further methodological development.

To elucidate the shared contributions of LLM-generated embeddings, we conduct an exploratory analysis on two groups of drug–cell line combinations. The first group consists of combinations for which all LLM-embedding–based models correctly predict the observed synergy effect, whereas the second group includes combinations for which all such models fail to do so. Importantly, the ground-truth synergy labels for both groups are positive, enabling a mechanism-level investigation into why LLM-derived embeddings succeed or fail in capturing synergistic interactions. Figure \ref{fig:syner pred}(c) presents a representative example from these two groups: the same drug pair, 5-FU and dasatinib, is applied to different target cell lines, yielding divergent prediction outcomes. 

Through consultation with expert chemists and biologists, we analyze these cases and derive mechanistic explanations at three complementary levels. In brief, VCaP cells \cite{ATCC_CRL2876} are driven by a dominant and well-defined signaling axis, most notably the androgen receptor (AR) pathway \cite{knuuttila2014castration}, that directly interacts with the targets of the administered drugs, thereby producing a relatively “clean” biological signal that is more amenable to computational prediction. In contrast, MSTO-211H cells \cite{ATCC_CRL2081} exhibit biphasic growth patterns and pronounced cellular and molecular heterogeneity, with complex and non-linear signaling networks that introduce substantial biological noise and hinder accurate prediction. At the phenotypic level, VCaP is largely composed of epithelial cells, whereas MSTO-211H contains heterogeneous cell populations; this difference in cellular homogeneity further favors predictive performance in VCaP. At the level of driver mechanisms, VCaP is strongly dependent on AR signaling. Dasatinib, a tyrosine kinase inhibitor \cite{mestermann2019tyrosine}, directly inhibits Src kinase (SRC) \cite{liu2010dasatinib}, which in turn modulates AR signaling in VCaP cells, yielding a coherent and biologically interpretable interaction. By contrast, MSTO-211H is characterized by mixed and less well-defined oncogenic drivers, making its regulatory landscape substantially more difficult to model. Finally, from a drug-target logic perspective, MSTO-211H likely engages multiple redundant or compensatory pathways, further reducing predictability. Therefore, one key insight we discovered is that the predictability of drug synergistic effects is closely linked to whether the description of the drugs involved is sufficiently clear and whether the target cell type can be accurately defined. One possible hypothesis at this stage is that incorporating more efficacy-related information into drug descriptions, such as EC values for individual drugs, may yield better prediction results in related downstream applications.

\textbf{Evaluations of LLM embeddings for drug-protein interaction prediction.} Drug-Protein Interactions (DPIs) refer to the binding of a drug molecule to a target protein, which is crucial for determining the drug's mode of action and fate in the living organism. Understanding DPIs is vital for drug development and personalized medicine \cite{ruffner2007human}, as these interactions can influence a drug's distribution, rate of excretion, and potential toxicity. Previous research has shown that using deep learning methods to produce drug and protein embeddings can help predict DPIs, while the embeddings from LLMs lacks similar functional exploration. To make a fair evaluation, we control the source of protein embeddings as ESMC \cite{hayes2025simulating}, which is an advanced embedding model that can encode protein sequences into vectors. We also select three datasets from a benchmark for DPI prediction of traditional methods (named based on the data sources, as Human, DrugBank, and C.elegans; \cite{tdckexinhuang}), and compare drug embeddings from different LLMs to assess their performances by using Accuracy (ACC) and AUROC (AUC) as metrics.

\begin{figure}[H]
    \centering
    \includegraphics[width=1\linewidth]{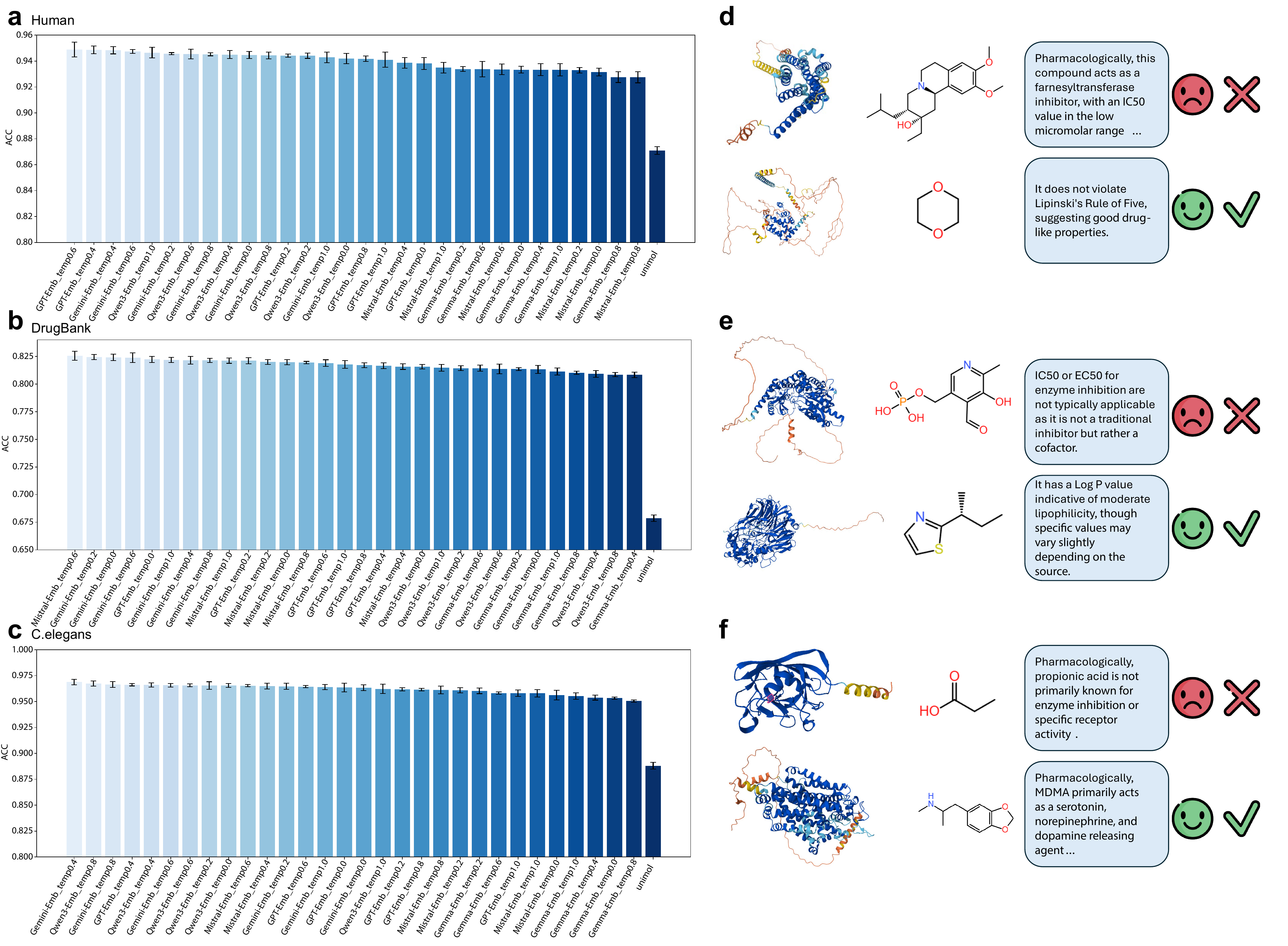}
    \caption{Benchmarking results for LLM embeddings of drug-protein interaction prediction. (a)-(c) represent accuracy based on the prediction results with different LLM-generated drug embeddings across three datasets. (b)-(d) represent the drug-protein pair with detected interaction (up) and no interaction (bottom) across three datasets. We also provide the textual description of drugs from GPT-4o.}
    \label{fig:dpi_pred}
\end{figure}

We train a multilayer perceptron (MLP) to predict drug–protein interactions from embedding representations and summarize overall performance using accuracy (ACC) in Figures \ref{fig:dpi_pred} (a)–(c) across the three benchmark datasets. Overall, LLM-generated drug embeddings consistently outperform domain-specific embeddings, and embeddings derived from Gemini models rank highly across leaderboards on multiple datasets. A similar trend is observed for AUROC (AUC), as shown in Supplementary Figures \ref{supfig:dpi_auc} (a)–(c). We further observe that generating drug descriptions at relatively higher temperatures can improve predictive performance in this task, likely because increased textual diversity enhances the likelihood of capturing protein-relevant functional information. Notably, no single LLM embedding model consistently ranks first across all datasets, indicating dataset-specific strengths. GPT-derived embeddings exhibit clear advantages for predicting human drug–protein interactions, whereas embeddings from Mistral and Gemini models perform better on curated knowledge-base datasets such as DrugBank. In contrast, Gemini and Qwen3 embeddings are more effective for DPI prediction in C. elegans. 

We further investigate whether DPIs can be inferred from LLMs and how LLM-derived embeddings contribute to DPI prediction. To this end, we select representative samples from all three datasets, including both experimentally validated interactions and confirmed non-interactions, for qualitative analysis. As illustrated in Figures \ref{fig:dpi_pred} (d)–(f), drugs annotated as having DPIs are typically associated with more definitive and informative textual descriptions, such as explicit statements of drug-likeness, known targets, or mechanism-of-action details. In contrast, compounds without recorded DPIs often exhibit substantially higher uncertainty in their descriptions: some are not clearly characterized as drugs, while others contain only partial or ambiguous information about pharmacological properties. These observations suggest that LLM-derived drug embeddings effectively encode probabilistic cues embedded in natural language descriptions and map them into a continuous vector space that is informative for interaction prediction. By contrast, models trained solely on molecular structures lack access to this contextual and semantic information, limiting their ability to capture drug–protein relationships and resulting in inferior predictive performance. Taken together, these results highlight the importance of aligning embedding choices with dataset characteristics and provide practical guidance on selecting LLM-derived embeddings for DPI  prediction.

\textbf{Evaluations of drug embeddings for chemical perturbation prediction.} Directly characterizing the effects of drugs on cellular activity is essential for elucidating mechanisms of action and potential side effects, thereby enabling the development of more effective therapeutics. In this section, we integrate single-cell RNA sequencing (scRNA-seq) with drug perturbation data to predict drug-induced gene expression changes at the cellular level, facilitating validation and interrogation of drug targets. This paradigm is commonly referred to as chemical perturbation sequencing \cite{peidli2024scperturb,stathias2020lincs}. Prior studies have shown that predicting the perturbation effects of unseen drugs remains highly challenging for machine-learning–based models, largely because training data cannot exhaustively cover all drug conditions and test sets often involve severe out-of-distribution scenarios. To address this challenge, previous work has introduced general-purpose drug representations, such as low-dimensional molecular descriptors derived from RDKit \cite{bento2020open} or embeddings from MFMs \cite{ji2024uni}. Given that LLMs can also encode rich drug-related information from text, we explore the use of LLM-derived embeddings as alternative representations for chemical perturbation prediction.

To this end, we leverage a large-scale transcriptomic dataset with chemical perturbation annotations, Tahoe 100M \cite{zhang2025tahoe}, which comprises over 100 million transcriptomic profiles measuring the effects of approximately 1,100 small-molecule perturbations across 50 cancer cell lines. Tahoe 100M contains 14 plates, and thus we select one plate for validation, one plate for testing, and the rest of plates are used for training. To overcome the challenges introduced by data noise, we aggregate the single-cell-level transcriptomic profiles with the related covariates (cell line, chemical perturbation, and dosage), and this pre-processing method is also introduced by the original manuscript for downstream analysis. The  task is to predict the perturbed gene expression profiles based on the input chemical information and control-level gene expression profiles. Our selected methods are ChemCPA \cite{lotfollahi2023predicting}, which is a well-known domain-expert method for predicting the perturbation effect with the help of drug embeddings from RDKit (as the baseline embeddings, ChemCPA\_baseline), and the selected metrics are R2, which are computed with different numbers (10, 20, and 50) of differential expression genes (DEGs) as well as all genes. To collect drug embeddings, we consider using LLM-generated embeddings with different embedding models from the descriptions produced by GPT-4o under different temperatures (0-1.0).

\begin{figure}[H]
    \centering
    \includegraphics[width=1\linewidth]{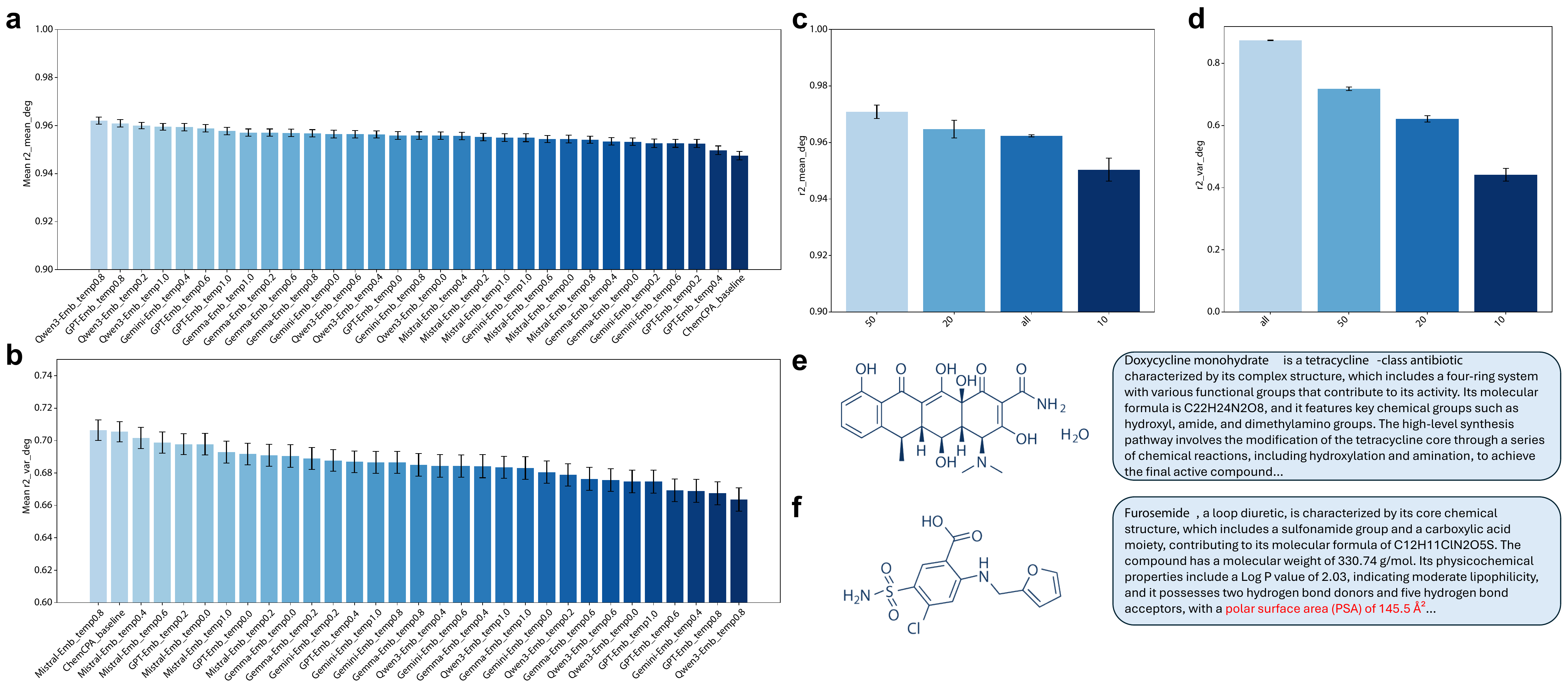}
    \caption{Benchmarking results of LLM-generated embeddings to predict drug perturbations. (a) The comparison of average R2 ($R^2$) computed with DEG set across different LLM-generated drug embeddings. (b) The comparison of R2 variance computed with DEG set across different LLM-generated drug embeddings. (c) The comparison of average R2 across different sides of genes. (d) The comparison of R2 variance across different sides of genes. (e) Structure and text description of the molecule whose perturbation has the highest prediction accuracy. (f) Structure and text description of the molecule whose perturbation has the lowest prediction accuracy.}
    \label{fig:pert_model}
\end{figure}

We first visualize the averaged R2 and the corresponding variance across different methods in Figures \ref{fig:pert_model} (a), (b) and Supplementary Figure \ref{supfig:lfc_deg}, accordingly. Figure \ref{fig:pert_model}(a) demonstrates that incorporating drug embeddings derived from LLMs consistently improves predictive performance relative to the baseline approach, providing clear evidence of the added value of LLM-generated representations. Among all configurations, the combination of Qwen3-Emb with drug descriptions generated by GPT-4o at a temperature of 0.4 achieves the highest average R2 while maintaining low variance across different gene set sizes and drug categories. In contrast, the combination of Mistral-Emb with GPT-4o–generated descriptions at a temperature of 0.8 exhibits the largest variance, underscoring the importance of selecting an appropriate LLM embedding model and description-generation strategy.

We further conduct a sensitivity analysis with respect to the number of genes used for evaluation, as shown in Figures \ref{fig:pert_model} (c) and (d), which report the corresponding average R2 and variance. Using an intermediate number of genes (e.g., 50) yields the highest overall R2, whereas restricting the analysis to a small gene set (e.g., 10 genes) introduces substantially greater variability. Computing R2 over all genes results in the largest variance, which is expected given the heterogeneous difficulty of predicting perturbation effects across different drugs.

Finally, Figures \ref{fig:pert_model} (e) and (f) present the structural and textual characteristics of the drugs with the highest and lowest R2 values under the best-performing configuration. Our analysis reveals that the description of doxycycline monohydrate contains richer biologically informative annotations, such as its classification as a tetracycline antibiotic, which facilitates accurate prediction of gene-level effects. In contrast, the description of furosemide emphasizes physicochemical properties with limited biological context, weakening the linkage between the drug and downstream gene regulation. These observations suggest that combining robust embedding models with biologically informative textual descriptions is critical for achieving strong performance in multimodal tasks involving complex biological data.

\section{Discussion and Conclusion}
Advances in LLMs have accelerated scientific discovery, yet systematic evaluation frameworks remain lacking in specialized subfields such as at the intersection of biology and chemistry that defines drug discovery. Therefore, in this study, we investigate the capacities of using LLMs to generate textual descriptions as well as representations of chemical molecules, and further evaluate their abilities in handling domain-specific applications across three critical scenarios of drug synergy, DPI, and drug perturbation predictions. Our work not only summarizes current advances in LLMs for drug research and identifies challenges, but also reveals the specific domains where different LLM models excel. In collaboration with domain experts, we have designed a set of practical guidelines with clear directional insights that are both timely and highly valuable.

Our evaluations for the quality of LLMs-generated drug descriptions show that GPT-series models still have the leading positions in formulating the annotations of drugs in the same context, and accurately summarize most of the important properties based on their knowledge-based memory. At the same time, we observe that adjusting the prompt for the query model still impacts the results. Notably, after using a Meta prompt modification to reorient the LLM from a general domain to the chemical domain, the quality of the model's responses showed a significant improvement. Other hyper parameters, such as temperature and random seed do not obviously alter the quality of the LLM output. Therefore, the knowledge background within LLMs can help us standardize drug descriptions, thereby aligning different researchers' understanding of the same drug and accelerating communication and collaboration.

In regards the evaluations of LLM-generated drug representations, using LLM embedding models always outperform domain-specific models. However, we found that the ranks of LLMs' embeddings are highly task-specific, and even embeddings from different models might have strong similarity (such as embeddings from GPT-Emb and Gemini-Emb). The best LLM embedding model used for drug synergy prediction is from the Gemini series, although GPT-Emb, Mistral-Emb, and Qwen3-Emb models all show good performances in predicting the drug-protein interaction. To predict the chemical perturbation effect under the gene expression level, both Qwen3-Emb and Mistral-Emb models show leading performances, but we also find that the variance of prediction might be increased by applying LLM-generated embeddings for this task. Meanwhile, through analyzing the prediction results, we found that we can examine the impact of input descriptions on downstream tasks by examining the differences in model performance. For example, in the DPI prediction task, descriptions of drug samples labeled as having no interaction relationship typically lack specific drug-likeness metrics and clinical evidence compared to those with such relationships. Therefore, integrating LLMs for generating drug descriptions with drug representations into a unified pipeline may represent a promising future direction for AI-accelerated drug discovery.

It should be noted that our evaluation also identified numerous limitations in the current generation of LLM models. First, LLMs may generate factually incorrect descriptions, such as specific molecular weights. Although this information does not fall under functional descriptions, it still negatively impacts the reliability of LLMs. Second, we found that LLMs struggle to accurately generate chemical structural information, such as two-dimensional molecular structures. Therefore, future efforts should incorporate structural information into model training to unify the structure-function-property framework. 

Finally, we provide guidance on the application of LLMs in drug discovery. First, GPT-based models with chemical-centralized system prompt remain the preferred choice for generating drug descriptions and can identify functionally relevant types with high accuracy. Second, for LLM-generated embeddings, the Gemini-series models and advanced open-source embedding models are also viable tools. The selection of specific models requires balancing multiple requirements, including accuracy, cost, and runtime. 

\section{Methods}

\textbf{Dataset construction.} For drug function analysis, we collect a molecular description dataset from MolTextNet \cite{zhu2025moltextnet}, which contains two million molecule-text pairs and the chemical properties of selected molecules are well annotated in this dataset. For drug-target prediction, we consider three datasets from TDC \cite{tdckexinhuang}. Each dataset contains an indicator for a combination with drug and protein to represent the existence of interaction between proteins and drugs. For the drug synergy prediction task, we select D1 \cite{preuer2018deepsynergy}, D2 \cite{preuer2018deepsynergy}, and D3\cite{wang2022deepdds} datasets from BAITSAO \cite{liu2025building}, which are widely used baselines for drug synergy, and cover the results of synergy computed with different rules to further guarantee generalization. For the perturbation prediction task, we consider one dataset from ChemCPA \cite{hetzel2022predicting} and one dataset from Tahoe 100M \cite{zhang2025tahoe}. These datasets contain different drugs and phenotype information (e.g., cell types, disease states, etc.) and can measure the performance of models in predicting the perturbation effect of drugs on cellular activities.

\vspace{2mm}
\noindent 
\textbf{Details of content-level evaluation.} In order to evaluate each LLMs' performance in generating drug descriptions, we first create a sample of 1000 drugs randomly selected from the MolTextNet dataset \cite{zhu2025moltextnet}. We then extracted their common names from PubChem and filtered those out without common names, creating a final sample dataset of 862 drugs.   The input ``drug\_name" for each cancer drug is its common name extracted from the PubChem according to its SMILES.

In this work we investigated three types of prompts including normal or standard, chain of thought (CoT), and meta-cognition (Meta) prompts. Table \ref{tab: prompts} shows the specific three prompt templates used in our experiment for description generation.  Table \ref{tab:descript-generation-model} lists all the models used for generating the text descriptions for the given cancer drugs. In addition to Claude-opus4-20250514, which is only used to test normal prompt, and Claude-sonnet-4, only used to test CoT and Meta prompts, the rest of the LLMs test the performance of all three prompts in drug description generation. Moreover, seed control was not available for DeepSeek-v3 and the Claude4 models, whereas for all other models that support the \textit{seed} parameter (42 as a fixed value). Finally, Table \ref{tab:parameter-setting} lists all the other parameter settings. We examined all the temperatures (0, 0.2, 0.4, 0.6, 0.8, 1.0) to observe the models' performance and consistency in description generation. Therefore, in this study, we test 90 combinations of model, temperature, and prompt types in total.  For each of the 90 combinations, we generated three samples per drug to test the consistency of LLMs.

\begin{table}[htbp]
\centering
\begin{tabular}{|c|p{10cm}|}
\hline
\textbf{Prompt Type} & \textbf{Content} \\
\hline
Normal &
f``Please describe the drug \{drug\_name\} including structure information and composition, property (such as physicochemical and acid-base properties), pharmacological activity, and synthesis in one paragraph. The physicochemical property can cover information such as molecular weight, Log P (lipophilicity), numbers of hydrogen bond donors/acceptors, polar surface area (PSA), and lipinski’s rule of five violations. As for the acid-base property, the value of pKa (acidic) and pKa (basic) can be covered. The pharmacological activity can include information such as IC50 (farnesyltransferase inhibition) and EC50 (Ha-Ras processing). About the synthesis, information such as SCS (synthetic complexity score), SAS (synthetic accessibility score), and synthesis pathway can be offered." \\ \hline
CoT &  f``Perform a step-by-step analysis of the drug \{drug\_name\} and synthesize all key information into one comprehensive paragraph. Clearly demonstrate your reasoning process: Frist, determine the core chemical structure (e.g., backbone, key functional groups) and molecular composition (elements, atom count). Calculate or provide its molecular weight. Second, based on Step 1, derive or retrieve: Log P (lipophilicity), Number of hydrogen bond donors (HBD) and acceptors (HBA), Polar surface area (PSA), Lipinski’s Rule of Five compliance (list violations, if any). Third, identify ionizable groups (e.g., carboxylic acids, amines). Report key acidic pKa and basic pKa values (if applicable). Fourth, focus on its primary mechanism (e.g., farnesyltransferase inhibition). Include potency metrics: IC50 for enzymatic inhibition (e.g., farnesyltransferase) and EC50 for functional effects (e.g., Ha-Ras processing inhibition). Fifth, Discuss synthetic feasibility, including Synthetic Complexity Score (SCS), Synthetic Accessibility Score (SAS) and Outline key steps in its synthesis pathway. Finally, synthesize all data from Steps 1-5 into one paragraph."\\ \hline
Meta & f``````You are a pharmaceutical chemistry expert. You need to describe the following features of \{drug\_name\}:

1. Structure \& Composition: Core chemical features and molecular formula, etc. 

2. Physicochemical Properties: Molecular weight, Log P (lipophilicity), numbers of hydrogen bond donors/acceptors, polar surface area (PSA), Lipinski’s Rule of Five violations. 

3. Acid-Base Properties: Acidic pKa and basic pKa values (if applicable). 

4. Pharmacological Activity: Primary mechanism (e.g., enzyme inhibition) with quantitative metrics such as IC50 (farnesyltransferase inhibition) and EC50 (Ha-Ras processing). 

5. Synthesis: Synthetic Complexity Score (SCS), Synthetic Accessibility Score (SAS), and high-level synthesis pathway. 

The final output format should be a single paragraph integrating all elements instead of bullet points."""\\
\hline
\end{tabular}
\caption{Overview of the Three Prompt Templates Used}
\label{tab: prompts}
\end{table}

\begin{table}[h]
    \centering
    \begin{tabular}{clc}\toprule
         \textbf{Full Name}&   \textbf{Prompt Applied}&\textbf{Seed Control}\\\midrule
         GPT4o& 
     All&Yes\\
 DeepSeek-V3& All&No\\
 Gemini-1.5-pro& All&Yes\\
 Claude-opus4-20250514& Normal&No\\
 Claude-sonnet-4& CoT, Meta&No\\
 Mistral-large-2411& All&Yes\\ \bottomrule\end{tabular}
    \caption{Details for the five LLMs.}
    \label{tab:descript-generation-model}
\end{table}

\begin{table}[h]
    \centering
    \begin{tabular}{cc}\toprule
         \textbf{Parameter}&  \textbf{Value}\\\midrule
         temperature& 
    0, 0.2, 0.4, 0.6, 0.8, 1.0\\
 top p=1&1\\
 seed&42 (if applicable)\\
 max output tokens&512\\
 presence penalty&0.0\\
 frequency penalty&0.0\\ \bottomrule\end{tabular}
    \caption{Parameter setting for the 5 LLMs}
    \label{tab:parameter-setting}
\end{table}

To measure the similarity in the language level of drugs, we consider the set of scores including BLEU, ROUGE-1, ROUGE-2, ROUGE-L, and BERT \cite{papineni2002bleu,lin2004rouge,zhangbertscore,jain1radgraph,miscllmreviewer}. The previous four metrics can assess the token-level evaluation, while the last metric can assess the semantic-level evaluation. Specifically, we computed the metric scores by comparing the generated descriptions and ground truth extracted from the curated sample dataset. Since we generated three sample descriptions for each drug under the identical setting, the final scores for each combination of model and temperature are the mean of the average metric scores of three samples for all drugs under the corresponding parameter setting.

\vspace{2mm}
\noindent
\textbf{Details of embedding-level evaluation.}
To compare different embedding models at the representation level, we evaluated five embedding models by using them to generate embeddings for all drug descriptions produced by the best-performing model–prompt–temperature combination. Such settings can be identified in the prior section on content-level evaluation. For each drug, three independently generated embeddings were available; these sample embeddings were averaged to obtain a single, stable embedding per drug.

For each embedding model, we computed drug-level cosine similarities between the generated description embeddings and the corresponding ground-truth embeddings. Specifically, for each drug, the cosine similarity was calculated between its embedding derived from model-generated descriptions and the embedding obtained from the ground-truth description using the same embedding model. A single similarity vector was then obtained for each embedding model. Finally, to quantify the similarity between embedding models, we computed pairwise cosine similarity between these model-level vectors. This resulted in a model × model similarity matrix that captures the degree to which different embedding models induce better drug information.

\vspace{2mm}
\noindent
\textbf{Details of Task-Specific Evaluation.} We consider three domain-specific downstream applications to assess the performances of LLM-generated embeddings for addressing practical problems. For the drug synergy prediction task, we leverage LLM-generated drug embeddings to predict the synergy effect under specific cell lines. The synergy effect is defined by binary labels (0 (No) and 1 (Yes)) and we measure the AUROC and Accuracy \cite{pedregosa2011scikit} between the predicted probability and observed labels. To evaluate the dataset with synergy effect as continuous variables, we computed the PCC and R2 \cite{pedregosa2011scikit} between the predicted value and observed value, for which higher metrics mean better embeddings. For the drug-protein interaction prediction task, we leverage LLM-generated drug embeddings paired with ESM-generated protein \cite{hayes2025simulating} embeddings to predict their interactions under different sources. Since this is a classification problem, we considered measuring the AUROC and Accuracy between the predicted probability and observed labels. Finally, fFor the chemical perturbation prediction task, we leverage LLM-generated drug embeddings to predict the gene expression profiles of selected cell types under the effect of applying this drug. Since this is a regression problem, we consider measuring the R2 \cite{pedregosa2011scikit} between the predicted expression profiles and observed expression profiles. We consider the measurement across a different number of selected genes. Higher metrics mean better embeddings. 

\vspace{2mm}
\noindent
\textbf{Computational cost and model access.} Our maximum running time for each dataset is 24 hours. For open-source models, we use 50GB MEM and an A100 GPU (80GB) core. To benchmark closed-source models, we purchase the API to access them.

\section{Code availability}  
The codes of \method{} can be found in \url{https://github.com/HelloWorldLTY/drugplayground} with MIT license.

\section{Data availability} 
All data are available online. We have summarized the download links of each dataset in Supplementary File 2. We have cited all used datasets in this manuscript.

\section{Acknowledgments}
This project is supported in part by OpenAI Research Access Program and Google Cloud Research Access Program. We used resources from the Yale High Performance Center (Yale HPC) to conduct all of the experiments. KS and THG thank the National Institute of Allergy and Infectious Disease grant U19-AI171954 for support.

\section{Author contributions}

T.L. designed the study with S.J. T.L., S.J., and F.Z. ran all the experiments. T.L., S.J., and K.S. performed analyses. All authors wrote the manuscript. H.Z. supervised this work.

\bibliography{sn-bibliography}

\newpage

\appendix
\counterwithin{figure}{section}
\renewcommand{\figurename}{Supplementary Fig.}
\renewcommand\thefigure{\arabic{figure}}  

\counterwithin{table}{section}
\renewcommand{\tablename}{Supplementary Tab.}
\renewcommand\thetable{\arabic{table}}


\section{Supplementary Figures}

\begin{figure}[H]
    \centering
    \adjustbox{center}{
    \includegraphics[width=1\linewidth]{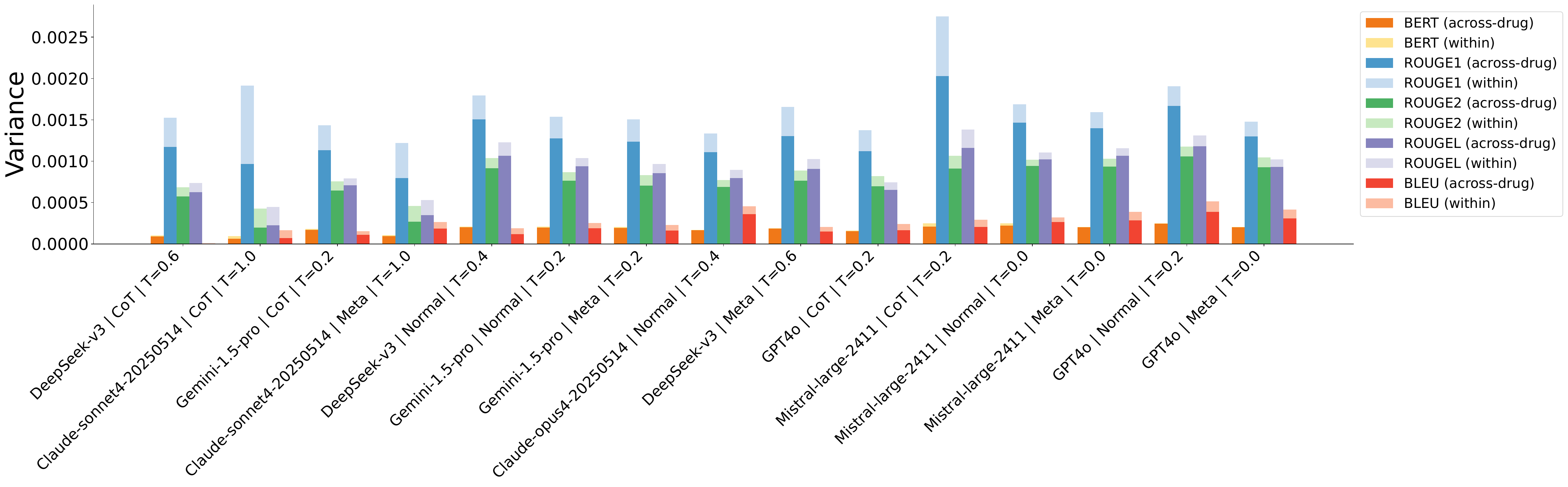}}
    \caption{Stacked Across-drug and Within-Variance over Metrics for the Top 15 Model–Prompt–Temperature Combinations}
    \label{fig: variances}
\end{figure}

\newpage 

\begin{figure}[H]
    \centering

    \begin{subfigure}{0.45\linewidth}
        \centering
        \includegraphics[width=\linewidth]{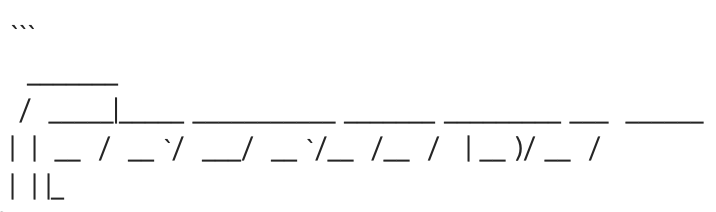}
        \caption{Failure example 1}
        \label{fig:ex1}
    \end{subfigure}
    \hfill
    \begin{subfigure}{0.45\linewidth}
        \centering
        \includegraphics[width=\linewidth]{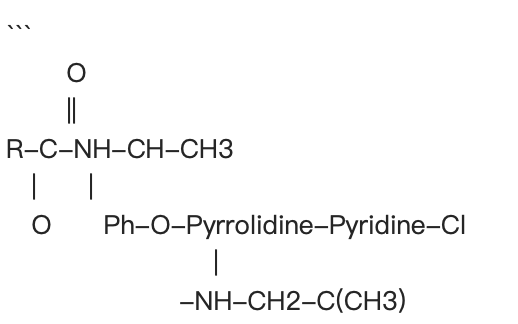}
        \caption{Failure example 2}
        \label{fig:ex2}
    \end{subfigure}

    \vspace{0.5em}

    \begin{subfigure}{0.3\linewidth}
        \centering
        \includegraphics[width=\linewidth]{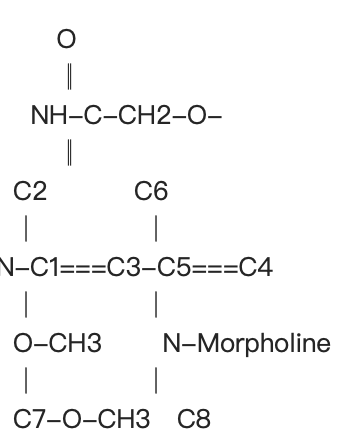}
        \caption{Failure example 3}
        \label{fig:ex3}
    \end{subfigure}
    \hfill
    \begin{subfigure}{0.4\linewidth}
        \centering
        \includegraphics[width=\linewidth]{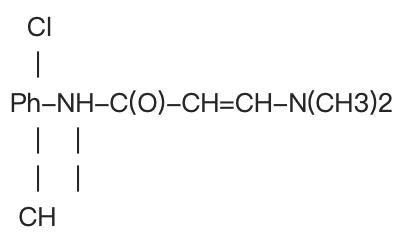}
        \caption{Failure example 4}
        \label{fig:ex4}
    \end{subfigure}

    \caption{Failure Examples of Molecule Structure in the Outputs of Mistral-large-2411 under standard prompts.}
    \label{fig:failure_examples}
\end{figure}


\begin{figure}
    \centering
    \includegraphics[width=1\linewidth]{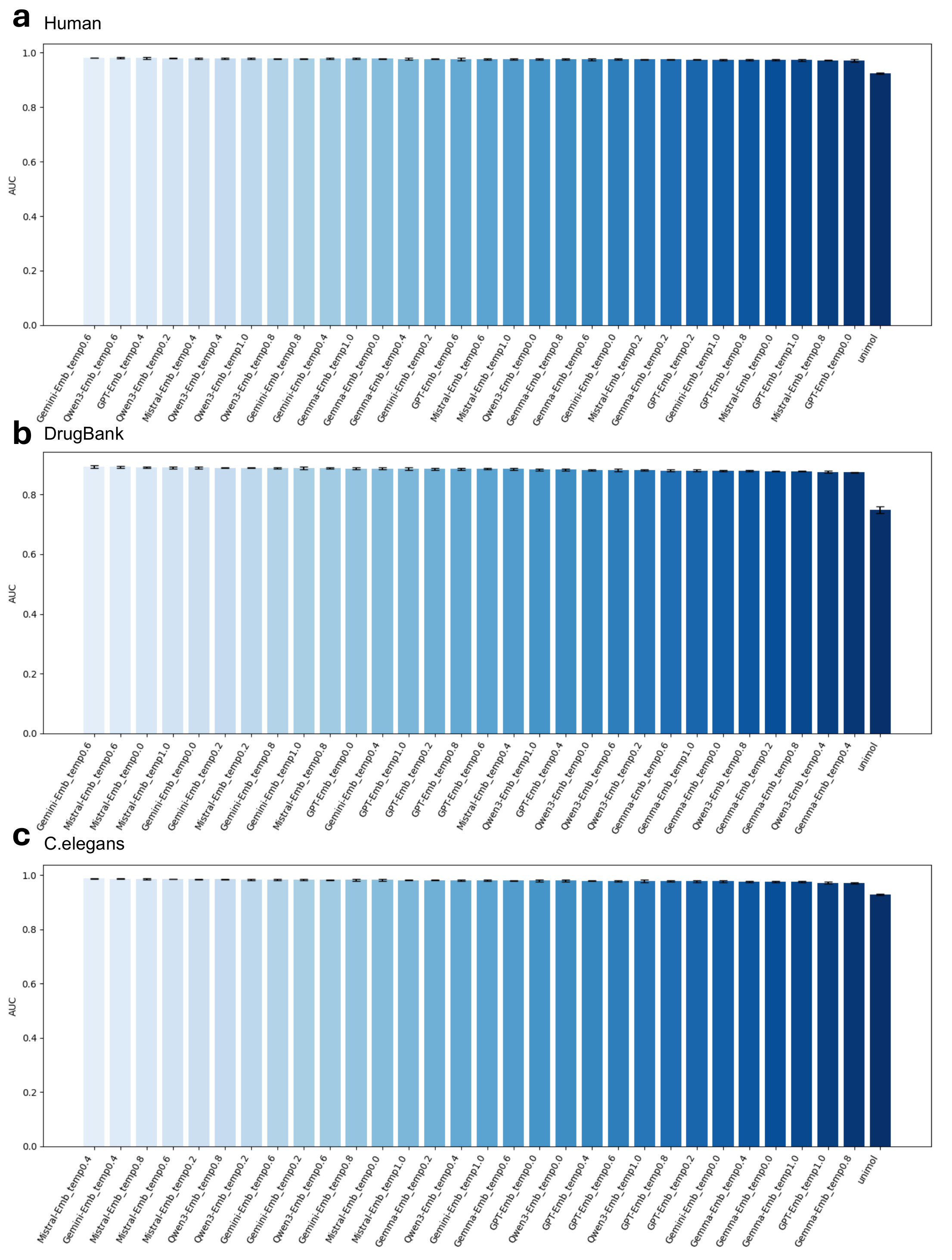}
    \caption{AUC of DPI prediction. (a)-(c) correspond to the scores from three different datasets.}
    \label{supfig:dpi_auc}
\end{figure}

\begin{figure}
    \centering
    \includegraphics[width=1\linewidth]{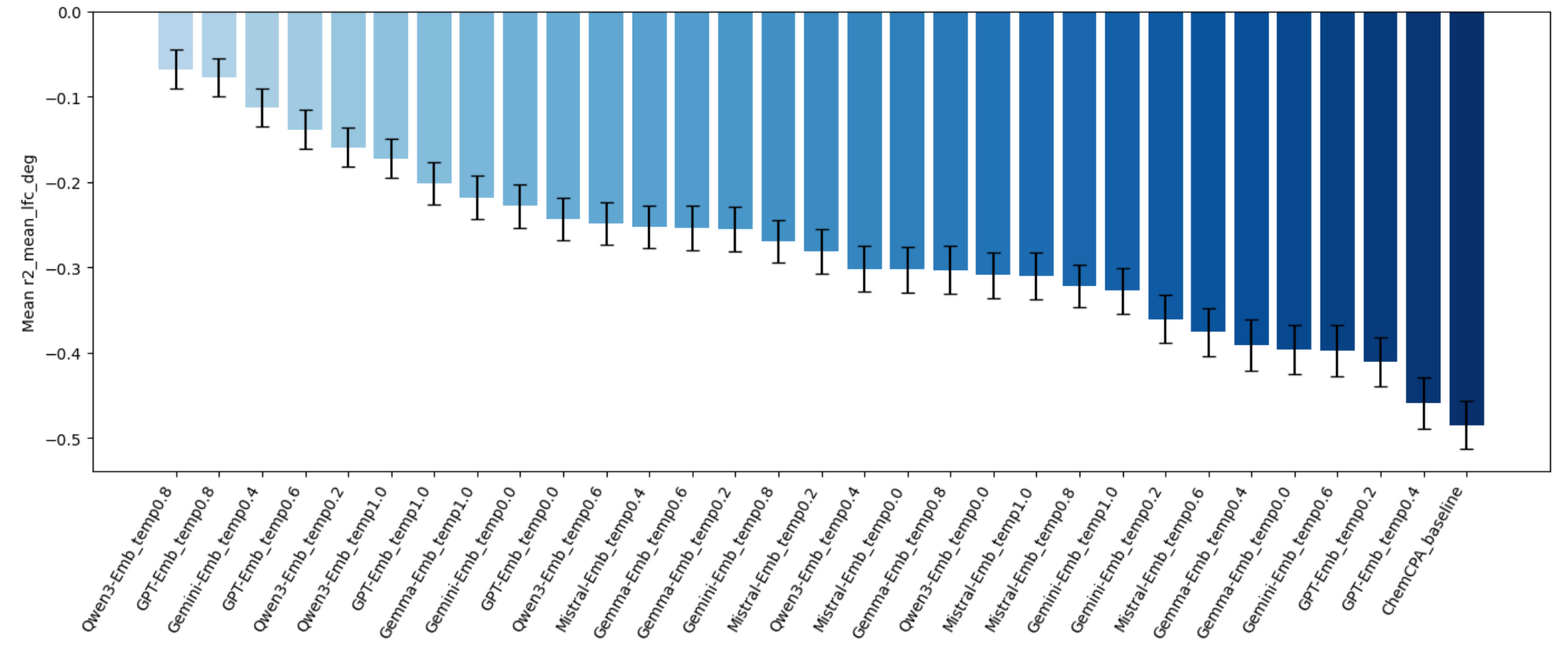}
    \caption{Average predicted log fold change based on drug embeddings from different models.}
    \label{supfig:lfc_deg}
\end{figure}

\clearpage
\section{Supplementary Tables}

\begin{table}[h]
\centering
\adjustbox{center}{
\begin{tabular}{>{\raggedright\arraybackslash}p{1cm}p{1cm}>{\centering\arraybackslash}p{0.7cm}>{\centering\arraybackslash}p{0.7cm}>{\centering\arraybackslash}p{1.59cm}>{\centering\arraybackslash}p{1.59cm}>{\centering\arraybackslash}p{1.59cm}>{\centering\arraybackslash}p{0.7cm}c}\toprule      
\textbf{Model}& \textbf{Prompt Type}& \textbf{Temp}& \textbf{BERT}& \textbf{ROUGE-1}& \textbf{ROUGE-2}& \textbf{ROUGE-L}& \textbf{BLEU}& \textbf{Avg Norm.}\\\hline
 GPT4o& Normal& 0.2& 0.8392& 0.4401& 0.1358& 0.2221& 0.0500&4.7698\\
 DeepSeek-v3& Normal& 0.4& 0.8284& 0.3762& 0.1113& 0.1966& 0.0231&3.0758\\\midrule
GPT4o& Meta& 0.0& 0.8407& 0.4622& 0.1425& 0.2189& 0.0468& 4.893\\
 DeepSeek-v3& CoT& 1.0& 0.7918 & 0.2752 & 0.0651 & 0.1287 & 0.0038 &0.1226 \\ \bottomrule

\end{tabular}
}
\caption{\textbf{Quantitative Performance Comparison Across Models, Prompting Strategies, and Temperatures} The Avg Norm. in the Table refers to the average normalized metric scores for the corresponding combination. The rest are defined as the same as above.}
\label{tab: worst-vs-best}
\end{table}

\begin{table}[h]
    \centering
    \begin{tabular}{cp{10.5cm}}\toprule
         \textbf{Type}&  \textbf{Example}\\\midrule
         \textbf{Inconsistency}&  \textbf{Expected Output}: Same structure and content of ground truth, such as: Chloroaustralasine C, with the canonical SMILES string ..., exhibits a molecular weight of $<$number$>$419.91$<$/number$>$ ...\\ 
         & \textbf{Example 1}: GPT4o | Meta | T0.0\\
         & \textbf{Actual inconsistent output 1}: Chloroaustralasine C is a complex natural product ...  includes a chlorinated indole alkaloid moiety. Its molecular formula is C22H19ClN2O3, ... \\
         & \textbf{Actual inconsistent output 2}: Chloroaustralasine C is a complex natural product ... includes multiple aromatic rings and chlorine substituents, contributing to its molecular formula. The physicochemical properties...\\
         
         & \textbf{Example 2}: DeepSeek-V3 | CoT | T1.0\\ 
         & \textbf{Actual inconsistent output 1}: \\
 &\#\#\# Step-by-Step Analysis of **2-[Ditert-butyl(fluoro)silyl]ethanol**  \\
 &...\\
 &- **Molecular formula**: $C_{10}H_{21}FO_{2}Si$ (10 carbons, 21 hydrogens, 1 fluorine, 2 oxygens, 1 silicon)...\\
         & \textbf{Actual inconsistent output 2}: \\
 &\#\#\# Step-by-Step Analysis of 2-[Ditert-butyl(fluoro)silyl]ethanol:\\
 &...\\
         & - **Molecular Formula**: $C_{10}H_{23}FO_{2}Si$ (derived from: 2 t-Bu groups = 2 × $C_4H_9$ = $C_8H_{18}$, plus Si, F, and $–CH_2CH_2OH$ = $C_2H_5O_2$)...\\
 \hline
 \textbf{Missing}&\textbf{Expected Output}: Give specific values of the corresponding characteristics required in the prompt, such as The most acidic pKa is $<$number$>$12.45$<$/number$>$\\
 &\textbf{Example}: GPT4o | Meta | T0.0\\
 &\textbf{Output Missing Info}: ... include a basic pKa around 12 for the guanidino group of arginine, while the acidic pKa values are less prominent due to the peptide nature. ...\\

\hline         
\textbf{Truncation}&  \textbf{Expected Output}: Complete description containing at least all the elements mentioned in the prompt.\\
 &\textbf{Example 1}: Claude-sonnet4-20250514 | Meta | T0.2\\
 &\textbf{Truncated Output 1}: ... requiring a multi-step convergent approach that typically involves the stereoselective formation of the 1\\
 &\textbf{Example 2}: Mistral-large-2411 | CoT | T0.0\\
 &\textbf{Actual Truncated Output 2}: **Step-by-Step Analysis of N-Methyl-N'-1,3-thiazol-2-ylurea** ... **Step 3: Ionizable Groups and p\\

\hline
\textbf{Hallucinations}&  \textbf{Expected Output}: a paragraph directly introduce the characteristics of the given drug without adding unnecessary message.\\
 &\textbf{Example 1}: DeepSeek-V3 | CoT | T0.4\\ 
 &\textbf{Hallucinated Output 1}: \#\#\# Step-by-Step Analysis of 5-Benzyl-9-bromopaullone:

--- \#\#\#\# **Step 1: ...\\
 &\textbf{Example 2}: Mistral-large-2411 | CoT | T0.0\\
 &\textbf{Hallucinated Output 2}: To analyze the drug 4-(2-Methyl-7-phenylthieno[3,2-d]pyrimidin-4-yl)morpholine, we start by ... Next, we derive or retrieve key physicochemical properties...\\
 &\textbf{Example 3}: Gemini-1.5-pro | Normal | T0.6\\
 &\textbf{Hallucinated Output 3}: ... The proposed synthesis pathway is a general suggestion and specific reaction conditions would need to be optimized.\\
 \hline
    \end{tabular}
    \caption{Examples of Inconsistent, Truncated, Hallucinated Generated Descriptions}
    \label{tab: ITH}
\end{table}

\end{document}